\documentclass[letterpaper]{article}

\usepackage{natbib,alifeconf}  
\usepackage{amsmath}
\usepackage{amssymb}
\usepackage{url}
\usepackage{booktabs}
\usepackage{graphicx}     
\usepackage{subcaption}   
\usepackage{caption}      

\title{Wrong Face, Wrong Move: The Social Dynamics of Emotion Misperception in Agent-Based Models
}

\author{David Freire-Obreg\'on $^{1}$ \\
\mbox{}\\
$^1$SIANI, Universidad de Las Palmas de Gran Canaria, Spain \\
david.freire@ulpgc.es} 

\begin{document}
\maketitle

\begin{abstract}
The ability of humans to detect and respond to others' emotions is fundamental to understanding social behavior. Here, agents are instantiated with emotion classifiers of varying accuracy to study the impact of perceptual accuracy on emergent emotional and spatial behavior. Agents are visually represented with face photos from the KDEF database and endowed with one of three classifiers trained on the JAFFE (poor), CK+ (medium), or KDEF (high) datasets. Agents communicate locally on a 2D toroidal lattice, perceiving neighbors' emotional state based on their classifier and responding with movement toward perceived positive emotions and away from perceived negative emotions. Note that the agents respond to perceived, instead of ground-truth, emotions, introducing systematic misperception and frustration.
A battery of experiments is carried out on homogeneous and heterogeneous populations and scenarios with repeated emotional shocks. Results show that low-accuracy classifiers on the part of the agent reliably result in diminished trust, emotional disintegration into sadness, and disordered social organization. By contrast, the agent that develops high accuracy develops hardy emotional clusters and resilience to emotional disruptions. Even in emotionally neutral scenarios, misperception is enough to generate segregation and disintegration of cohesion. These findings underscore the fact that biases or imprecision in emotion recognition may significantly warp social processes and disrupt emotional integration.

\end{abstract}

\begin{figure}[ht]
    \centering
    \includegraphics[width=0.8\columnwidth]{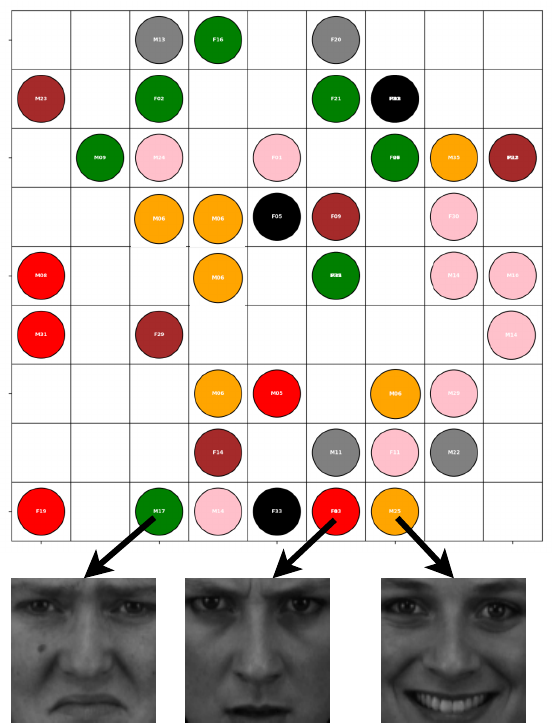}
    \caption{Lattice configuration of emotional agents of the simulation. Each agent is represented as a colored circle, where color encodes the agent's current emotional state (e.g., happy, sad, angry, etc.).}
    \label{fig:intro}
\end{figure}

\section{Introduction}
Recognizing and responding to the emotional states of others is a fundamental component of human social interaction. Successful emotion perception promotes trust, cooperation, and group cohesion, and misperceiving emotions leads to misunderstandings, conflict, or social exclusion. Artificial entities and social robots that increasingly interact with humans and with other robots need the ability to perceive and respond to emotional information.

Studies on emotional contagion in social networks have long interested social and psychological researchers~\citep{hatfield1993emotional,barsade2002ripple}. The rapid explosion of the Internet and, in particular, the spreading of social network websites exposed the possibility of the transmission of emotional states through virtual contacts as well~\citep{bollen2011happiness,kramer2012spread}. For instance, ~\cite{kramer2014experimental} conducted a large-scale controlled experiment on Facebook and uncovered that emotional displays on posts influence others' emotions even without direct contact. In the same tradition of work, ~\cite{Ferrara15} also investigated the emotional tone of messages before people's tweets and uncovered apparent emotional influence patterns.

Out of such findings, several computational models have been proposed to capture the temporal dynamics of emotional contagion~\citep{bosse2015agent,wang2015esis}. To generalize the susceptible--infected--susceptible (SIS) model so that it can accommodate spontaneous emotional adoption, ~\cite{hill2010emotions} proposed the so-called SISa model in this regard. To capture the integration of emotion weighting into the transmission probability of the message so that contagious emotions impact the dissemination process of information, ~\cite{wang2015esis} proposed the ESIS model. In reality, however, most models assume the existence of perfect or homogeneous emotional perception skills of the agents and do not take into consideration the variability and imperfection present in human and artificial recognition systems.

This assumption also introduces a critical flaw: emotional understanding is never absolute, even in humans, and certainly not in computer programs. Classifiers trained from limited or culture-bound datasets typically generalize poorly even in the usage environment. These defects can deflect an agent's comprehension of others' feelings and thus influence its decisions about approaching, avoiding, or mismatching with peers.
An agent-based modeling (ABM) framework is presented in which each agent possesses a vision system with distinctive perceptual accuracy given by a convolutional neural network (CNN) trained on one of three facial expression datasets. Agents are depicted visually with images from the KDEF dataset and act locally on a 2D toroidal lattice. Their social behavior, given by movement towards or away from others, is governed not by the actual emotional state of neighbors but rather the agent's understanding of neighbors' emotions. This allows us to study the impact of systematic misperception on the emergence, breakdown, and stability of emotional and spatial patterns.

Through experiments with homogeneous and heterogeneous populations, we show that low-accuracy classifiers distort emotions, reduce trust, amplify negative convergence (especially sadness), and lead to social breakdown. In contrast, perceptually accurate agents foster stability and harmonious clusters. Perceptual biases (arising from data, culture, or technology) critically affect social structure and emotional health. Our model, a stylized abstraction, illustrates how systematic misrecognition of emotions can destabilize cooperative societies.

\section{Background}
Trust and cooperation are the basis of human social life, and the ability to recognize the trustworthiness of others is thought to be a key evolutionary advantage in the growth of cooperative behavior \citep{boone2003emotional}. Facial signals long thought to be connected with physiological arousal can be relatively reliable indicators of internal states connected with reliability \citep{SCHUG201087}. Those who freely express emotion are typically judged to be more cooperative and more trustworthy \citep{lount2010impact}. Even though considerable work has been conducted looking at trust from the perspective of rational behavior and economic decision-making, much less focus has been placed on the role of emotional expression in the establishment of interpersonal trust \citep{lount2010impact}. Combining emotional and rational perspectives on trust may offer a more comprehensive understanding of the mechanics underlying social cooperation.

This fundamental function of emotional expression in social cognition has also aroused much interest in the recognition and interpretation of such expressions in humans and machine systems.

\textbf{Detection of Emotions in Humans and Machines}. Facial expression recognition has been deeply studied in psychology and computer science. We adopt the widely used Ekman set of six basic emotions plus neutral~\citep{ekman1993facial}, enabling comparability across JAFFE~\citep{lyons1998jaffe}, CK+\citep{lucey2010ckplus}, and KDEF\citep{lundqvist1998kdef}, datasets widely employed for training deep models~\citep{Li22,Salas-Caceres2024}. Although the models show very high accuracy on in-dataset testing, cross-dataset verification leads to drastically reduced performance~\citep{icaart20}, a sign of weak generality. More recent psychology work also calls into question the assumption that facial expressions map reliably to discrete emotions~\citep{barrett2012emotions}. These questions about the models' validity raise concerns about the practicality of using emotion AI in interactive systems. 

\textbf{Agent-Based Models (ABMs)}. ABMs have been popular tools for exploring complex social phenomena, including opinion formation, cooperation, and segregation. Historically, two principal limitations have plagued ABMs: computational inefficiencies on large scales and restricted behavioral sophistication~\citep{bonabeau2002agent}. More recent advancements in the domains of differential programming and deep learning have served to overthrow performance bottlenecks through the possibility of vectorized computation and neural-inspired models of large populations of agents~\citep{andelfinger2023towards}. Parallel efforts to elevate behavioral realism have aimed to leverage large language models (LLMs), drawing on the latter's human-like reasoning capabilities in order to model more subtle agent behavior~\citep{kerr2021covasim}. LLMs' principal disadvantage continues to be the computationally expensive cost of inference~\citep{vezhnevets2023generative}. Here, we present an alternative solution: behavioral expressiveness based on the use of various perceptual modeling, entirely avoiding language generation and still producing rich emergent behavior on the population level.

\section{Methodology}


We model a population of $N$ agents $\{a_1, a_2, \dots, a_N\}$ situated on a two-dimensional toroidal grid $\mathcal{G} \subset \mathbb{Z}^2$. Time evolves in discrete steps $t \in \mathbb{N}$. Each agent interacts locally with its Moore neighborhood and adjusts its internal state based on perceived emotional stimuli, classifier confidence, and accumulated trust.

\textbf{Emotion Classification via CNN.}
Each agent is equipped with a pretrained convolutional neural network (CNN) $f_\theta : \mathbb{R}^{H \times W \times C} \rightarrow \mathbb{R}^{|\mathcal{E}|}$ for facial emotion recognition. Given an input image $x \in \mathbb{R}^{H \times W \times C}$, the output is a vector of logits:
\[
f_\theta(x) = \left[ s_1, s_2, \dots, s_{|\mathcal{E}|} \right],
\]
Moreover, the predicted emotion label is obtained as
\[
\hat{e} = \arg\max_{i} \, s_i,
\]
Where $\hat{e} \in \mathcal{E}$ and the label set is defined as
\[
\mathcal{E} = \{\text{happy}, \text{sad}, \text{angry}, \text{fear}, \text{disgust}, \text{surprise}, \text{neutral}\}.
\]

Each model is trained using a cross-entropy loss over a labeled dataset:
\[
\mathcal{L}(\theta) = - \sum_{i=1}^{|\mathcal{E}|} y_i \log \left( \text{Softmax}(f_\theta(x))_i \right),
\]
Where $y_i$ is the one-hot encoded ground-truth emotion.

Three CNN classifiers, denoted $f^{(1)}_\theta, f^{(2)}_\theta, f^{(3)}_\theta$, are pretrained independently on the CK+, JAFFE, and KDEF facial expression datasets, respectively. Each agent is assigned one of these models, $f^{(k)}_\theta$, according to the experimental configuration defined for the simulation.

\textbf{Agent Specification.}
The tuple describes each agent $a_i$:
\[
a_i = \langle e_i(t), I_i, f^{(k)}_\theta, T_i(t), \mathcal{H}_i(t), \vec{p}_i(t) \rangle,
\]
Where:
\begin{itemize}
    \item $e_i(t) \in \mathcal{E}$ is the emotional state at time $t$,
    \item $I_i$ is the set of facial images for agent $a_i$, one per emotion in $\mathcal{E}$, taken from the KDEF dataset. At each timestep, the agent’s displayed image is updated according to its current emotional state $e_i(t)$,
    \item $f^{(k)}_\theta$ is its assigned classifier,
    \item $T_i(t) \in [0, 1]$ is the dynamic trust level,
    \item $\mathcal{H}_i(t) = [e_i(t-h), \dots, e_i(t-1)]$ is a history buffer of size $h$,
    \item $\vec{p}_i(t) \in \mathcal{G}$ is the spatial position.
\end{itemize}

\subsection{Interaction Dynamics}

At each timestep $t$, the agent performs the following sequence:

\paragraph{(1) Emotional Perception.}

For each neighbor $a_j \in \mathcal{N}_i(t)$, the agent perceives an emotion:
\[
\hat{e}_j^{(i)}(t) = f_\theta^{(k)}(I_j).
\]

\paragraph{(2) Trust Update.}

Let $\delta_j(t) = \mathbb{I}\left[ \hat{e}_j^{(i)}(t) \neq e_j(t) \right]$ be the prediction error. Perceptual reliability (hereafter referred to as trust for brevity) is updated using an exponential moving average:
\[
T_i(t+1) = (1 - \alpha) T_i(t) + \alpha (1 - \delta_j(t)),
\]
where $\alpha \in (0,1)$ is a smoothing parameter. Note that the comparison with ground-truth labels in $\delta_j(t)$ is an evaluation device in the simulation rather than an observable feature available to agents. Agents only act based on their perceptions, but the ground-truth reference allows us to quantify the effect of perceptual mismatches on emergent trust dynamics. This mechanism captures the idea that repeated misperceptions reduce the agent’s confidence in the reliability of emotional information from neighbors.

\paragraph{(3) Environmental Valence.}

We define sets of positive and negative emotions:
\[
\mathcal{E}_+ = \{\text{happy}, \text{surprise}, \text{neutral}\},
\]
\[
\mathcal{E}_- = \{\text{angry}, \text{sad}, \text{fear}, \text{disgust}\}.
\]

The perceived valence of the local environment is:
\[
V_i(t) = \sum_{j \in \mathcal{N}_i(t)} \left[ \mathbb{I}[\hat{e}_j^{(i)}(t) \in \mathcal{E}_+] - \mathbb{I}[\hat{e}_j^{(i)}(t) \in \mathcal{E}_-] \right].
\]
Agents enter an \emph{avoidance mode} whenever the perceived emotional valence falls below a threshold, that is, when
\[
V_i(t) < \tau_{\text{valence}}.
\]
In this state, the agent attempts to relocate to an adjacent unoccupied cell in the grid, effectively moving away from a negatively perceived neighborhood. Here only avoidance is implemented, but adding attraction to similar emotions (e.g., clustering negative states) would be a natural extension.

\paragraph{(4) Frustration-Based Emotion Switching}

If trust falls below certain thresholds, an agent may switch to a negative or confused state:
\[
e_i(t+1) = 
\begin{cases}
\text{sad}, & \text{if } T_i(t) < \tau_{\text{sad}} \text{ and } \text{sad} \in \mathcal{E}, \\
e_i(t), & \text{otherwise}.
\end{cases}
\]

We model sadness as the default frustrated state, as repeated social misperceptions typically elicit low affect and withdrawal rather than anger~\citep{carver2009}.

\paragraph{(5) Emotion Contagion}

Let $\mathcal{F}_i(t)$ denote the multiset of perceived emotional states in the 
neighborhood $\mathcal{N}_i(t)$ of agent $a_i$ at time $t$, that is,
\[
\mathcal{F}_i(t) = \{ \hat{e}_j^{(i)}(t) \mid a_j \in \mathcal{N}_i(t) \}.
\]
We define emotional contagion as occurring when a particular emotion 
$e^\star \in \mathcal{E}$ dominates the local neighborhood. Specifically, 
if there exists an emotion $e^\star$ such that
\[
\frac{\left| \{ a_j \in \mathcal{N}_i(t) \;|\; 
\hat{e}_j^{(i)}(t) = e^\star \} \right|}
     {\left| \mathcal{N}_i(t) \right|}
\;\geq\; \tau_{\text{contagion}},
\]
and $e^\star$ belongs to the agent's available expression set $\mathcal{E}$, 
then agent $a_i$ may adopt emotion $e^\star$ at the next timestep. 
This adoption is subject to a hysteresis condition: the candidate emotion 
$e^\star$ must not have appeared more than a fixed number of times in the 
agent's recent emotional history $\mathcal{H}_i(t)$, thereby preventing 
excessive emotional switching or oscillation.

\section{Experimental Setup}

The experimental setup consists of two main components: the training data for the emotion classifiers and the simulation parameters that define agent behavior and interaction dynamics. 

\textbf{Datasets.} The emotion classifiers were trained on three widely used facial expression datasets: JAFFE, CK+, and KDEF. The Japanese Female Facial Expression (JAFFE) dataset \citep{lyons1998jaffe} contains 213 grayscale images of 10 Japanese women posing six basic emotions (angry, disgust, fear, happy, sad, surprise) plus neutral. Its cultural and demographic homogeneity makes it a suitable test case for cross-population generalization. The Extended Cohn-Kanade (CK+) dataset \citep{lucey2010ckplus}, an extension of CK, includes 593 sequences and still images from 123 subjects (81\% Euro-American, 13\% Afro-American, 6\% other), annotated with eight categories: neutral, angry, contempt, disgust, fear, happy, sad, and surprise. Since \textit{contempt} is absent from JAFFE and KDEF, we omitted it to ensure consistency. The Karolinska Directed Emotional Faces (KDEF) dataset \citep{lundqvist1998kdef} provides high-resolution frontal face images of 70 actors under controlled conditions, each performing the seven basic emotions. For our experiments, we selected a balanced subset of 40 identities (20 male, 20 female). We extracted two frames per video (neutral and peak expression), yielding a balanced dataset across the six non-neutral emotions plus neutral.

\textbf{Implementation Details.} To test the model proposed above, we ran simulations on a controlled parameter and condition set. In our experimental scenario, we chose a population of $N = 40$ agents, matching the number of distinct identities used in the KDEF facial expression dataset, and every agent has the complete set of seven basic emotions. The simulation world is a toroidal grid of size $9 \times 9$ ($\mathcal{G} \subset \mathbb{Z}^2$), resulting in a total of 81 cells. With randomly located 40 agents, the design ensures roughly 50\% coverage of the surface of the grid. It provides an even spatial density with enough social contact coupled with room for agent movement and dispersal under the influence of emotional stress.  

These face images for emotion classification were preprocessed uniformly so that the images were resized to be $96 \times 96$ pixels in size, converted to grayscale, and normalized to the interval $[-1, 1]$, as is standard input to CNNs. The trust update mechanism uses a smoothing value of $\alpha = 0.1$, and this controls the rate at which the agents update the trust based on the accuracy of their perception. The agent also maintains a rolling buffer of emotional history of size $h = 5$, so that some history of the previous states influences its behavior now.

To control behavior in response to perceived emotional context, the valence threshold was set fixed as $\tau_{\text{valence}} = -1$, so the agents move when perceiving a strongly negative state. Change of emotions due to lacking trust is controlled based on a threshold of $\tau_{\text{sad}} = 0.3$, and emotional transmission via contagion requires that at least 70\% of the neighbors of an agent share a common emotion, i.e., $\tau_{\text{contagion}} = 0.7$. Each simulation ran for $T = 100$ discrete time-steps, and to make the results more robust, each parameter setting ran $R = 10$ times in order to capture stochastic variability from run to run.

\section{Experimental Evaluation}

To measure the influence of accuracy in emotional recognition on group emotional dynamics quantitatively, we model agent populations with varying perceptual skills. As previously specified, the agents perceive neighbors with the help of one of three trained emotion classifiers on data sets varying in generality to face images from KDEF. Agents change emotional state and spatial position based on perceived, not the ground-truth, emotions. Our design allows us to study the effects of misperception, trust decay, and affective feedback loops on group behavior and cohesion.

\subsection{Baseline: Emotional Recognition Accuracy}

Before analyzing emergent dynamics, we establish a baseline evaluation of the perceptual quality of each classifier in isolation. We test the JAFFE, CK+, and KDEF trained models on the full KDEF dataset (unseen during training for JAFFE and CK+) to simulate real-world generalization. This allows us to quantify classifier performance when interpreting agents represented by KDEF images, as used in all subsequent experiments.

Since all our simulation agents possess face images from the KDEF dataset used for visually representing all the agents in our simulations, we evaluated each classifier on this domain with the standard 80/20 train-test split. The directly KDEF-trained classifier performed pretty well with 96\% accuracy and macro-averaged precision and recall of 97\% and 96\%, respectively, and F1-score of 96\%, evidencing near-perfect training-test data alignment. The CK+ classifier, on the other hand, has minimal generalization capability and only achieves 37\% accuracy on KDEF with macro precision, recall, and F1-score of 36\%, 37\%, and 30\%, respectively. Worse performance still is observed for the classifier trained on the JAFFE dataset, which only achieves 19\% accuracy on KDEF and a macro F1-score of 10\%. The results indicate that the models of emotion recognition are highly dependent on the domain and demonstrate the way cultural and demographic differences between datasets can penalize cross-dataset generalization significantly. The all-Japanese female subject dataset JAFFE and the highly North American subject-based CK+ dataset equally struggle to interpret faces from the KDEF dataset, the latter having a more diverse collection of European faces. Simulation agents, then, with culturally mismatched datasets used for training, develop distorted concepts about the environment in which they exist and directly impact the social behavior and group emotional dynamics in the simulation.

\begin{figure}[ht]
    \centering
    \includegraphics[width=\columnwidth]{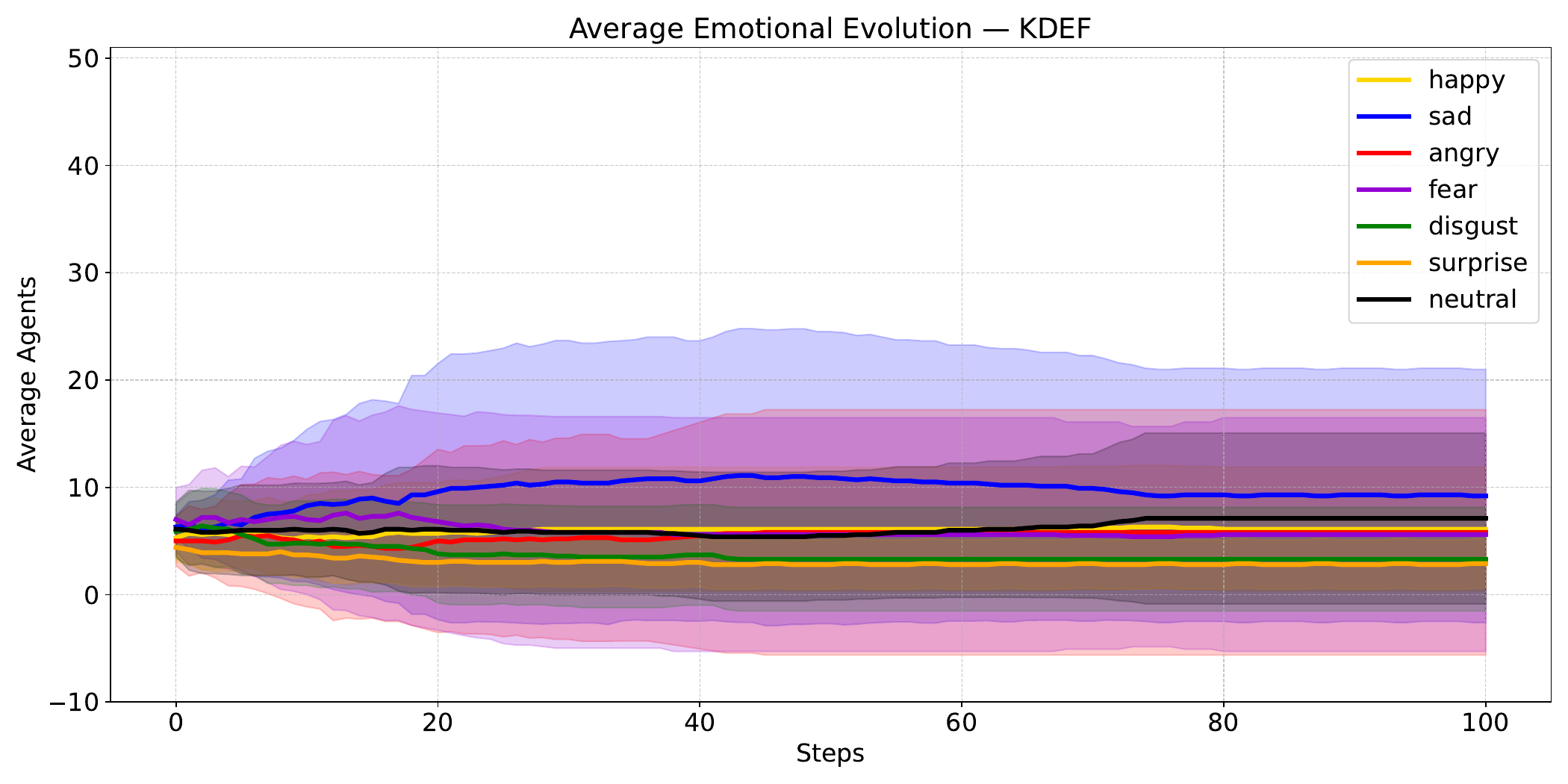}
    \caption{Emotional evolution in homogeneous agent populations using KDEF classifiers. Each plot shows the average emotion counts over 10 simulation runs.}
    \label{fig:exp1_kdef}
\end{figure}

\begin{figure}[ht]
    \centering
    \includegraphics[width=\columnwidth]{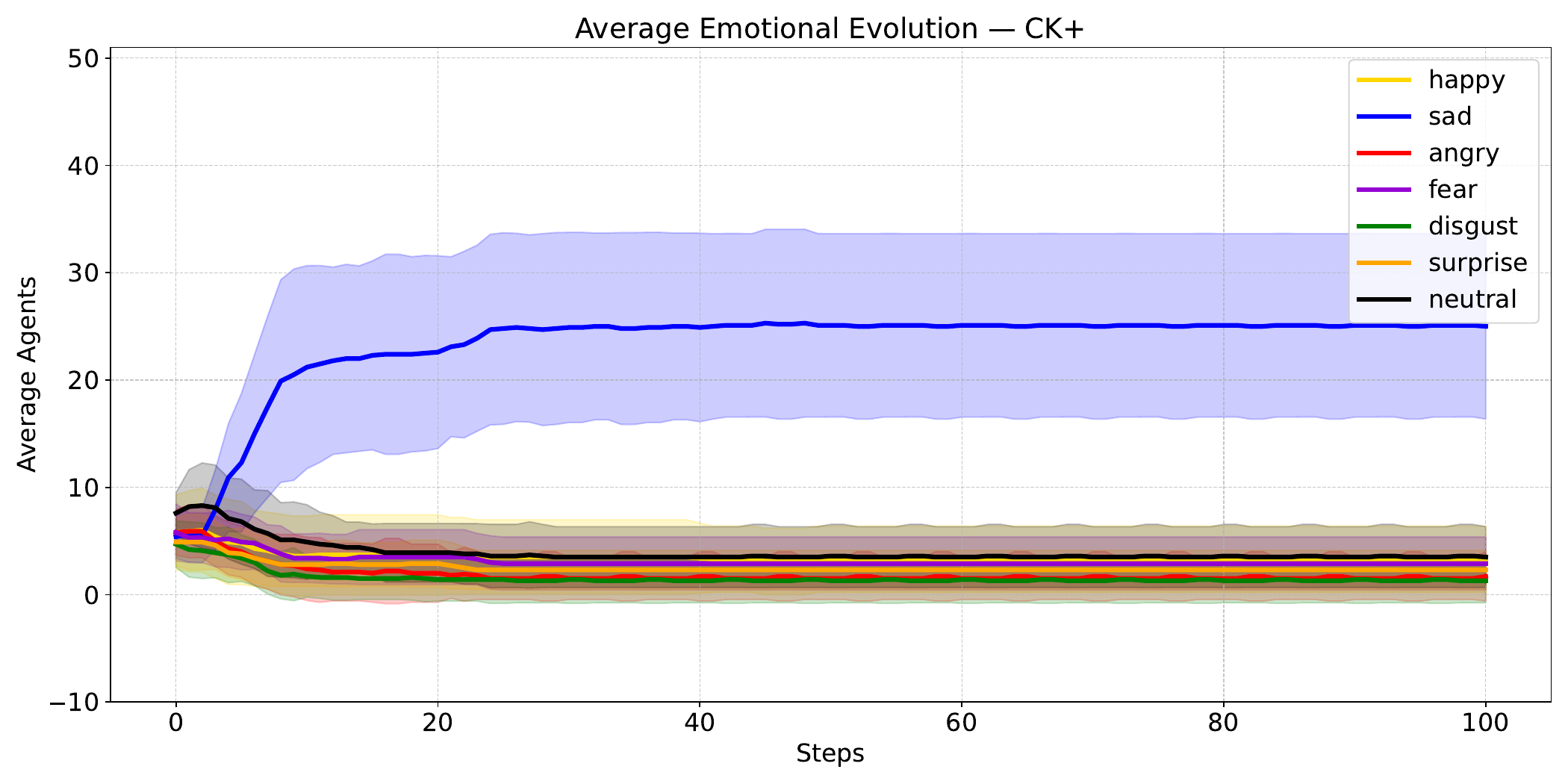}
    \caption{Emotional evolution in homogeneous agent populations using CK+ classifiers. Each plot shows the average emotion counts over 10 simulation runs.}
    \label{fig:exp1_ck}
\end{figure}

\begin{figure}[ht]
    \centering
    \includegraphics[width=\columnwidth]{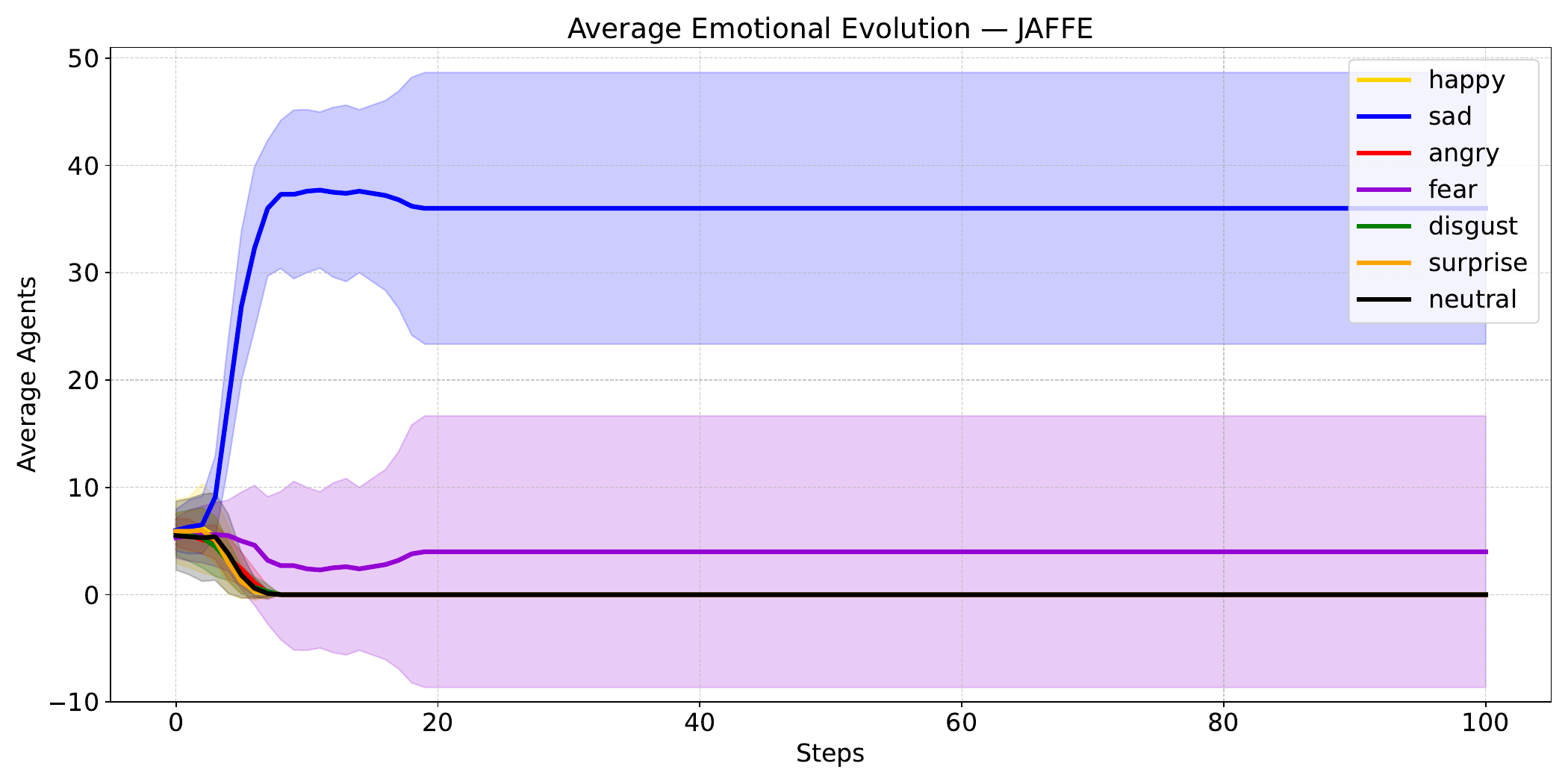}
    \caption{Emotional evolution in homogeneous agent populations using JAFFE classifiers. Each plot shows the average emotion counts over 10 simulation runs.}
    \label{fig:exp1_jaffe}
\end{figure}

\begin{table}[ht]
\centering
\caption{Cluster summary (average over final step) for individual datasets (KDEF, CK+, JAFFE) including the number of clusters, average cluster size, and maximum cluster size per emotion.}
\begin{tabular}{llccc}
\toprule
\textbf{Emotion} & \textbf{Condition} & \textit{Num} & \textit{Avg} & \textit{Max} \\
\midrule
Angry    & KDEF  & 5.80 & 1.63 & 2.60 \\
Disgust  & KDEF  & 4.60 & 1.16 & 1.40 \\
Fear     & KDEF  & 3.14 & 1.86 & 3.14 \\
Happy    & KDEF  & 2.29 & 4.17 & 6.29 \\
Neutral  & KDEF  & 3.62 & 2.45 & 3.75 \\
Sad      & KDEF  & 5.75 & 1.65 & 3.62 \\
Surprise & KDEF  & 2.25 & 1.79 & 2.12 \\
\midrule
Angry    & CK+   & 2.00 & 1.10 & 1.20 \\
Disgust  & CK+   & 2.00 & 1.05 & 1.20 \\
Fear     & CK+   & 2.38 & 1.19 & 1.38 \\
Happy    & CK+   & 2.67 & 2.39 & 3.17 \\
Neutral  & CK+   & 2.86 & 1.55 & 2.00 \\
Sad      & CK+   & 9.20 & 2.58 & 7.30 \\
Surprise & CK+   & 1.88 & 1.48 & 1.62 \\
\midrule
Fear     & JAFFE & 10.0 & 4.00 & 19.00 \\
Sad      & JAFFE & 10.0 & 4.19 & 16.56 \\
\bottomrule
\end{tabular}
\label{tab:cluster_individual}
\end{table}

\subsection{Experiment 1: Homogeneous Classifier Populations}

We start with entirely homogeneous populations
where all the agents equally rely on the same facial expression recognition model. Although the same initial state (forty agents,
random selection of emotional faces from the KDEF
dataset), the emergent classifiers produced drastically contrasting emergent group behavior. The disparities arise from the varying
perceptual accuracies of the models while interpreting facial
expressions from the KDEF faces.

When using only agents with the KDEF-trained classifier (see Figure \ref{fig:exp1_kdef}), the emotional landscape remains diverse and balanced throughout the simulation. No single emotion dominates: ``sad'', ``neutral'', ``happy'', and ``fear'' each appear with comparable frequency (between approximately 6 and 10 agents on average). Agent trust remains high at 0.96, reflecting accurate emotional perception and stable interactions. Clustering patterns are consistent with this emotional diversity: agents form moderately sized groups for most emotions, such as ``happy'' (average cluster size: 4.17) and ``neutral'' (2.45), with no overwhelming dominance or fragmentation.

The ABM with only the CK+ classifier yields a much more polarized outcome (see Figure \ref{fig:exp1_ck}). The ``sad'' emotion becomes dominant, accounting for nearly 60\% of the population on average (23.4 agents), while other emotions remain marginal. Trust drops significantly to 0.25, indicating frequent perceptual errors and agent frustration. This shift is also reflected in the cluster structure: although ``sad'' forms some larger clusters (up to 7.3 agents), other emotions tend to cluster weakly or not at all. The emotional drift toward sadness, despite a neutral or positive initialization, demonstrates how moderate perceptual inaccuracies can lead to unstable social dynamics.

The configuration with only the JAFFE-based agents displays the most extreme outcome (see Figure~\ref{fig:exp1_jaffe}). Nearly all agents converge to the ``sad'' state (average 34.7), with virtually no representation of other emotions. Trust falls to near zero (0.038), signaling near-constant misperception. The emotional homogeneity is reinforced by dense and large ``sad'' and ``fear'' clusters (maximum sizes over 16 agents), indicating runaway emotional contagion and the collapse of social differentiation.

These findings show that internal emotion models have a key role in determining emergent emotional and structural properties of agent populations. Clustering diversity and emotional stability are facilitated by high-fidelity classifiers (e.g., classifiers trained on the same data used for the visual inputs), and mismatched models create instability, miscommunication, and emotional polarization.
\begin{figure}[ht]
    \centering
    \includegraphics[width=\columnwidth]{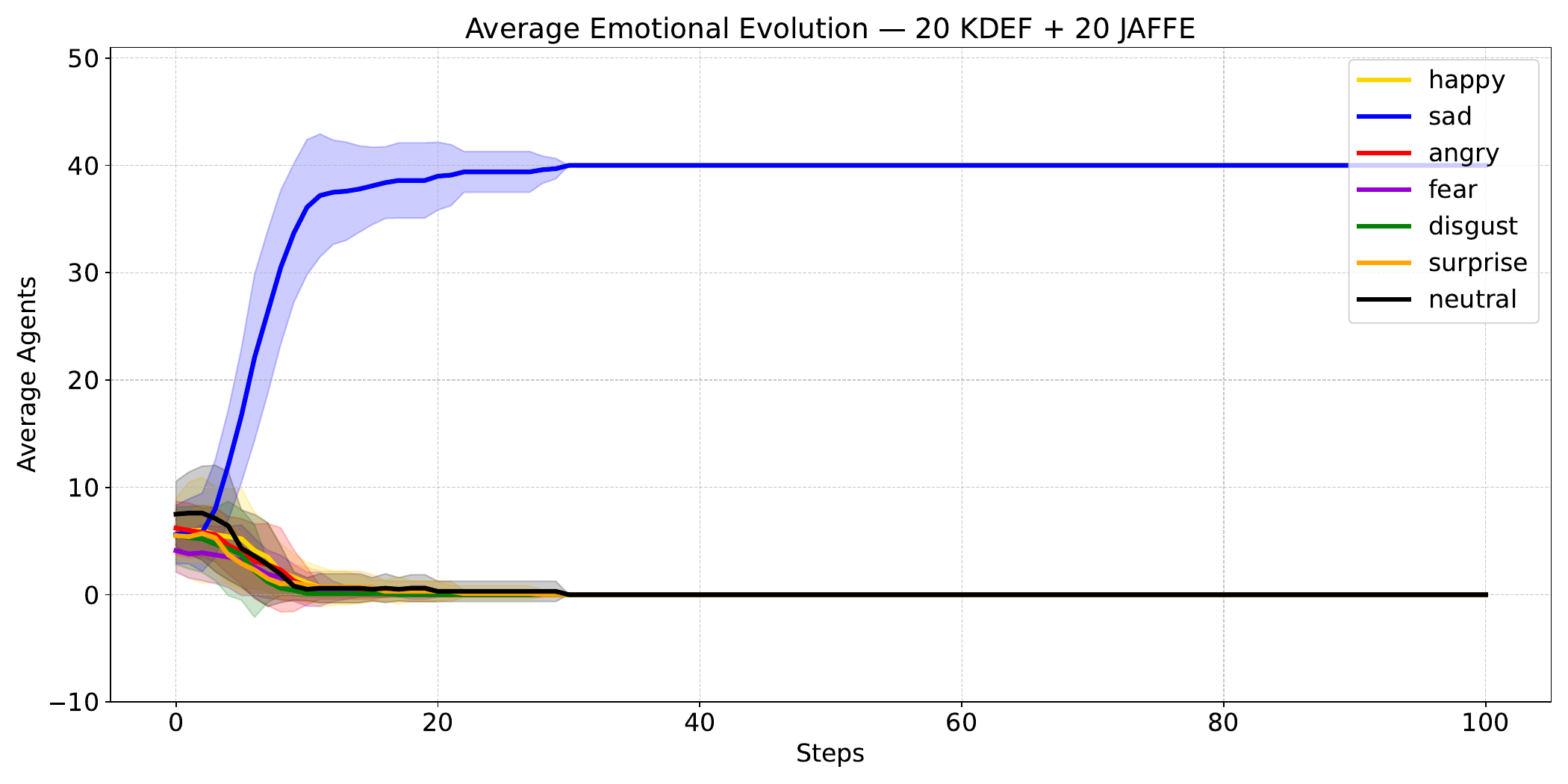}
    \caption{Emotional dynamics in mixed-agent populations with 20 KDEF and 20 JAFFE agents. Each plot shows average results across 10 simulation runs.}
    \label{fig:exp2_kdef_jaffe}
\end{figure}

\begin{figure}[ht]
    \centering
    \includegraphics[width=\columnwidth]{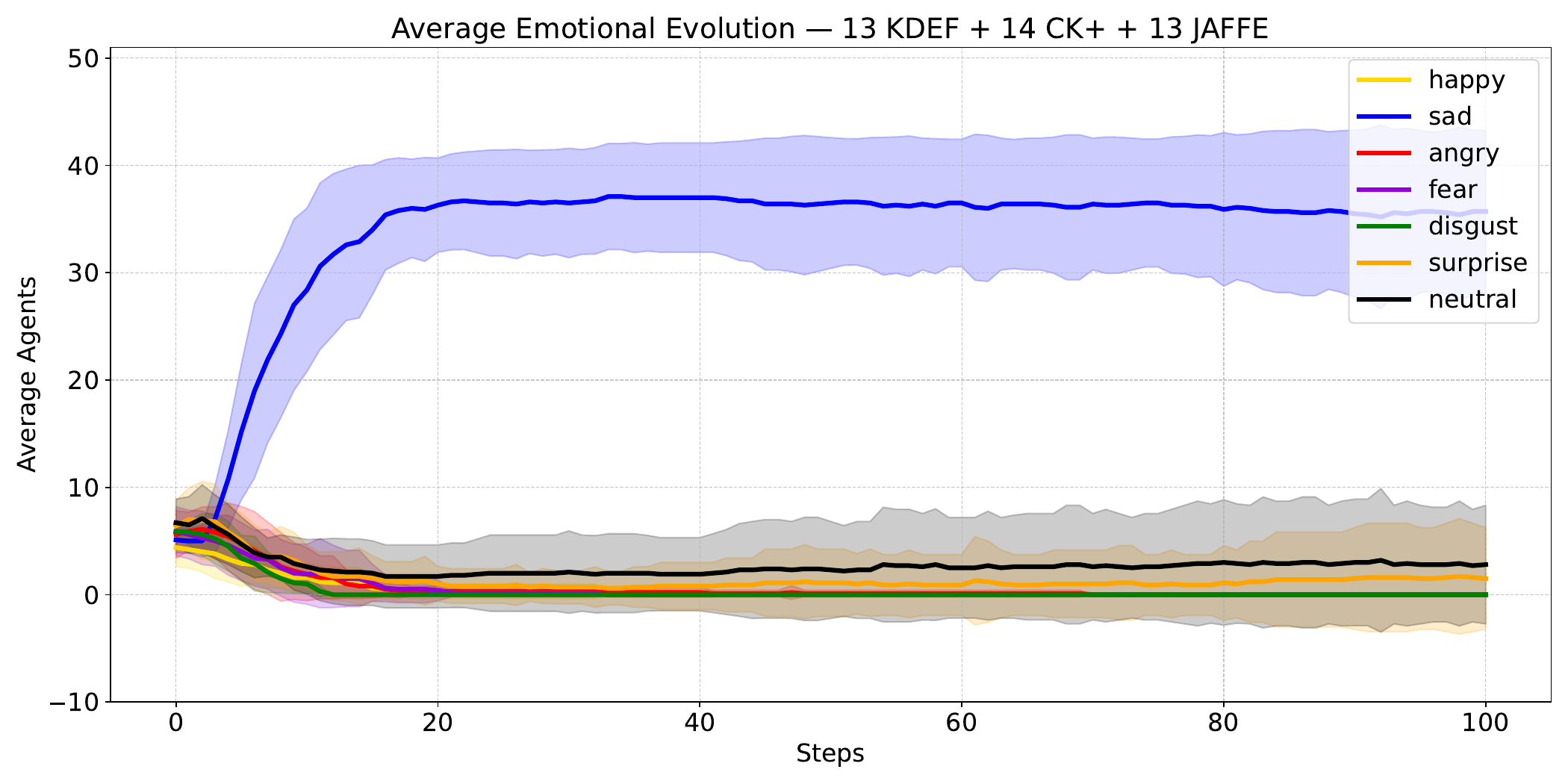}
    \caption{Emotional dynamics in mixed-agent populations with 13 KDEF, 14 CK+, and 13 JAFFE agents. Each plot shows average results across 10 simulation runs.}
    \label{fig:exp2_kdef_ck_jaffe}
\end{figure}

\begin{figure}[ht]
    \centering
    \includegraphics[width=\columnwidth]{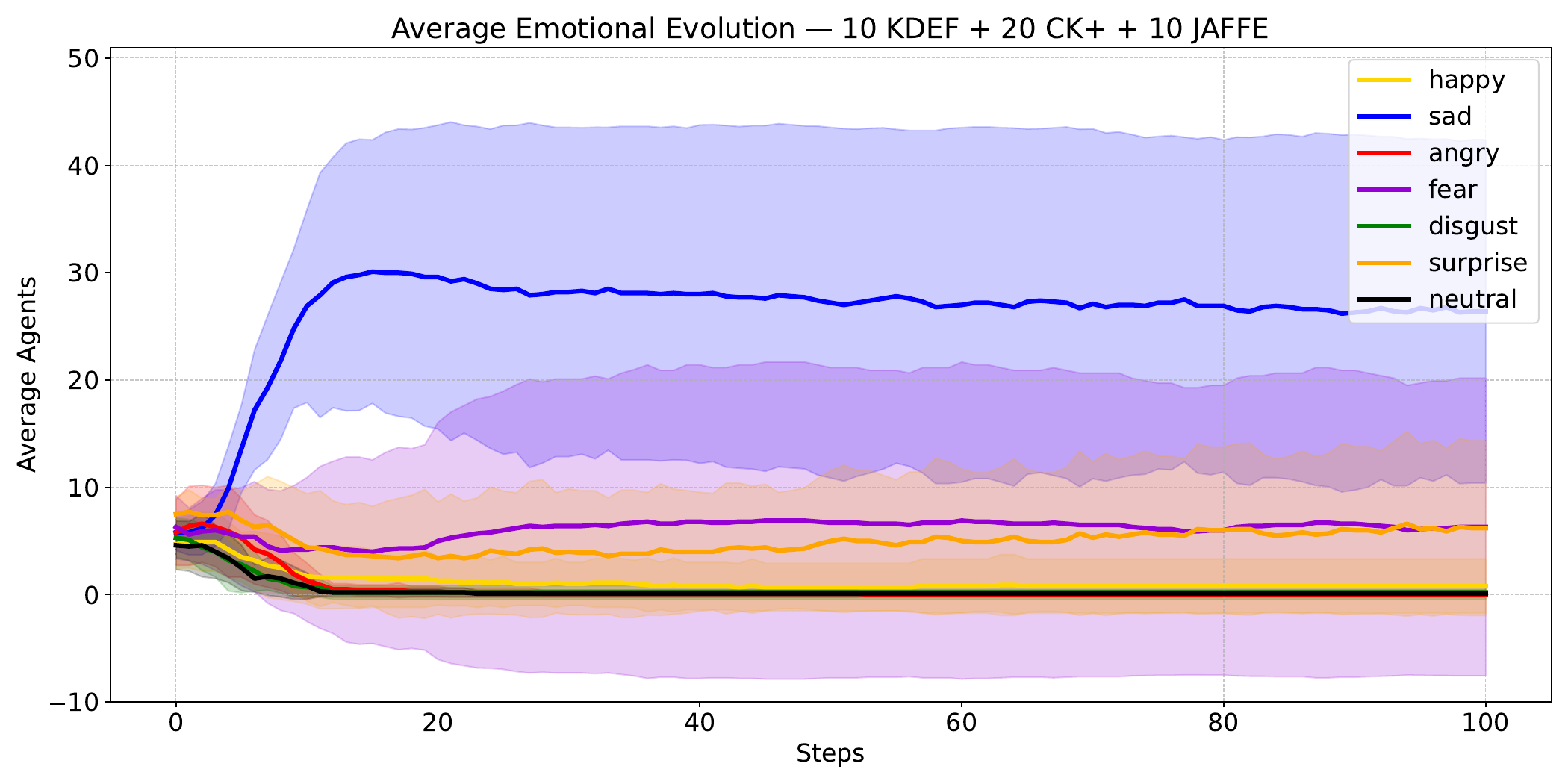}
    \caption{Emotional dynamics in mixed-agent populations with 10 KDEF, 20 CK+, and 10 JAFFE agents. Each plot shows average results across 10 simulation runs.}
    \label{fig:exp2_kdef_ck_jaffe_skewed}
\end{figure}

\begin{table}[ht]
\centering
\caption{Cluster summary (average over final step) for mixed agent populations, including the number of clusters, average cluster size, and maximum cluster size per emotion.}
\begin{tabular}{llccc}
\toprule
\textbf{Emotion} & \textbf{Condition} & \textit{Num} & \textit{Avg} & \textit{Max} \\
\midrule
Sad      & 20 KDEF + 20 JAFFE         & 9.50  & 4.42 & 15.50 \\
\midrule
Neutral  & 13 K + 14 C + 13 J         & 1.33  & 6.33 & 8.67  \\
Sad      & 13 K + 14 C + 13 J         & 10.50 & 3.61 & 13.00 \\
Surprise & 13 K + 14 C + 13 J         & 3.00  & 5.00 & 13.00 \\
\midrule
Disgust  & 10 K + 20 C + 10 J         & 1.00  & 1.00 & 1.00  \\
Fear     & 10 K + 20 C + 10 J         & 8.50  & 3.59 & 12.50 \\
Happy    & 10 K + 20 C + 10 J         & 1.00  & 7.00 & 7.00  \\
Neutral  & 10 K + 20 C + 10 J         & 1.00  & 1.00 & 1.00  \\
Sad      & 10 K + 20 C + 10 J         & 9.00  & 3.17 & 11.44 \\
Surprise & 10 K + 20 C + 10 J         & 2.00  & 7.62 & 13.25 \\
\bottomrule
\end{tabular}
\label{tab:cluster_mixed}
\end{table}

\subsection{Experiment 2: Mixed Classifier Populations}

To assess how agents with differing perceptual capacities influence each other, we simulated mixed populations consisting of multiple classifier types. Three specific configurations were tested: a balanced mix of 20 KDEF and 20 JAFFE agents; a uniform distribution of 13 KDEF, 14 CK+, and 13 JAFFE agents; and a population skewed toward CK+ with 10 KDEF, 20 CK+, and 10 JAFFE agents. All agents were initialized with random emotional expressions from the KDEF dataset, but the internal classifier used to interpret others' emotions varied by group.

In the \textbf{20 KDEF + 20 JAFFE} scenario (see Figure~\ref{fig:exp2_kdef_jaffe}), emotional outcomes are dominated by sadness, which accounts for more than 93\% of the population (mean = 37.4 out of 40). Other emotions are nearly absent, each averaging fewer than one agent. This result is accompanied by a complete collapse of trust among JAFFE agents (mean trust = 0.000), while KDEF agents maintain high confidence (trust = 0.945). The emotional clustering patterns reflect this imbalance, with large and dense clusters of ``sad'' agents (up to 15.5 agents), suggesting that the poor perceptual accuracy of JAFFE agents triggers a feedback loop of misinterpretation and contagion.

The \textbf{13 KDEF + 14 CK+ + 13 JAFFE} configuration also converges heavily toward ``sad'' (mean = 33.8), although with slightly more diversity than the previous case (see Figure~\ref{fig:exp2_kdef_ck_jaffe}). Emotions such as ``neutral'' and ``surprise'' appear sporadically. However, no emotion other than ``sad'' reaches a mean higher than three agents. Trust levels show a steep gradient: KDEF agents remain confident (0.951), CK+ agents drop to a modest level (0.116), and JAFFE agents again register no trust. Cluster structures reinforce this emotional dominance, with sadness forming the most significant and numerous clusters, although some presence of ``neutral'' and ``surprise'' groupings is also observed.

The most diverse pattern appears in the \textbf{10 KDEF + 20 CK+ + 10 JAFFE} condition  (see Figure~\ref{fig:exp2_kdef_ck_jaffe_skewed}). While ``sad'' still dominates (mean = 26.2), other negative emotions like ``fear'' (6.1) and ``surprise'' (5.0) also emerge more clearly. Small clusters of ``happy'' and ``disgust'' appear as well. Trust levels mirror classifier quality: KDEF remains high (0.948), CK+ moderate (0.223), and JAFFE low (0.063). Interestingly, this setting also yields a broader emotional distribution and more balanced cluster structures, ``happy'' and ``surprise'' form distinct and moderately sized groups (average cluster sizes of 7.0 and 7.6, respectively), suggesting that even limited perceptual improvement in the population can foster local zones of emotional variation.

Together, these findings demonstrate that population-level emotional dynamics are shaped not only by the overall perceptual quality of agents but also by their distribution within the social fabric. High-accuracy agents (KDEF) can maintain emotional stability in isolation or the presence of a minority. However, large proportions of low-performing classifiers (e.g., JAFFE) drive convergence toward negative affect and suppress emotional diversity across the population.

\begin{figure}[ht]
    \centering
    \includegraphics[width=\columnwidth]{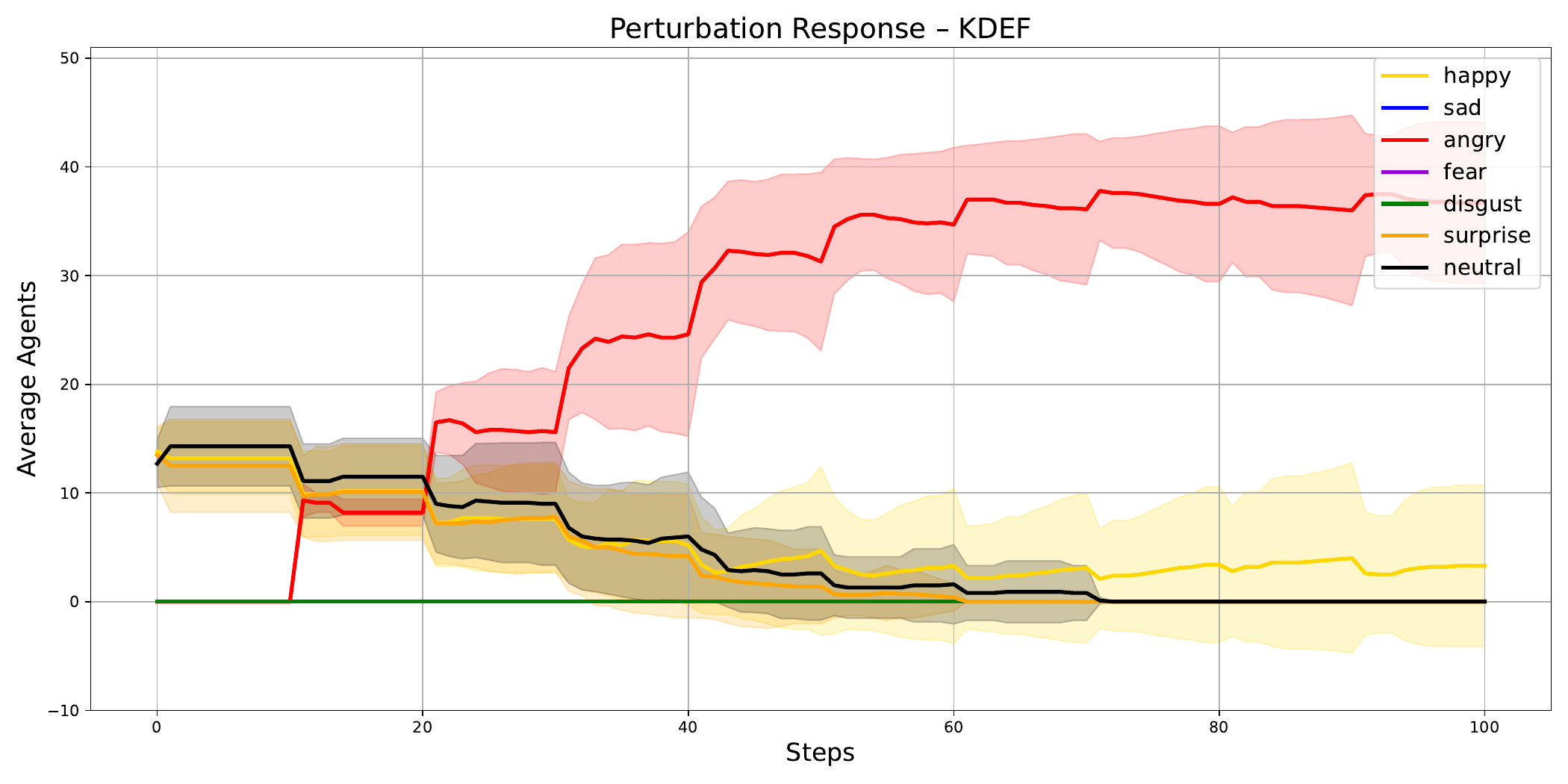}
    \caption{Emotional resilience of homogeneous KDEF agents under repeated emotional shocks. Each curve shows the decline of positive emotion ratio over time.}
    \label{fig:exp3_kdef}
\end{figure}

\begin{figure}[ht]
    \centering
    \includegraphics[width=\columnwidth]{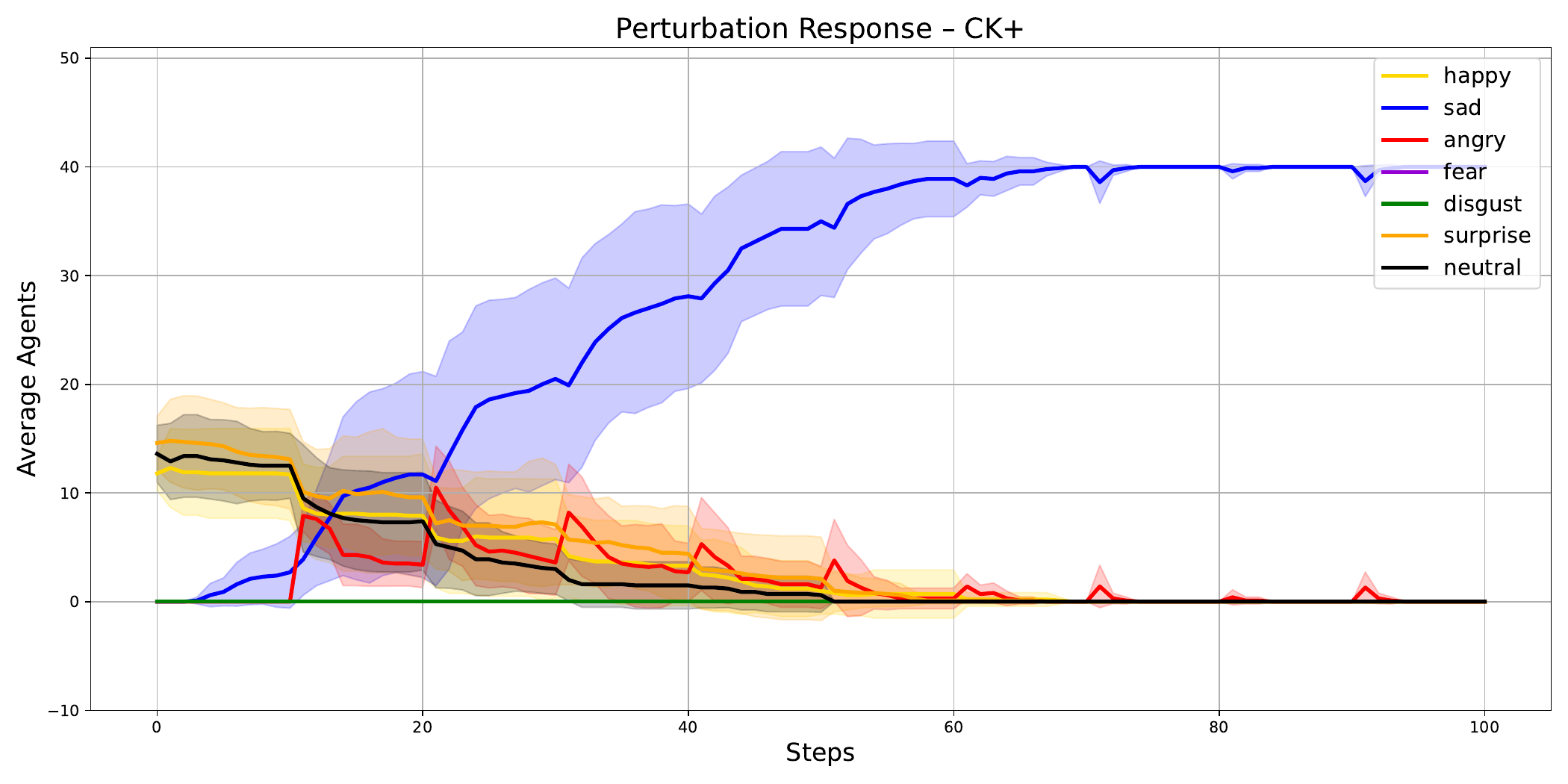}
    \caption{Emotional resilience of homogeneous CK+ agents under repeated emotional shocks. Each curve shows the decline of positive emotion ratio over time.}
    \label{fig:exp3_ck}
\end{figure}

\begin{figure}[ht]
    \centering
    \includegraphics[width=\columnwidth]{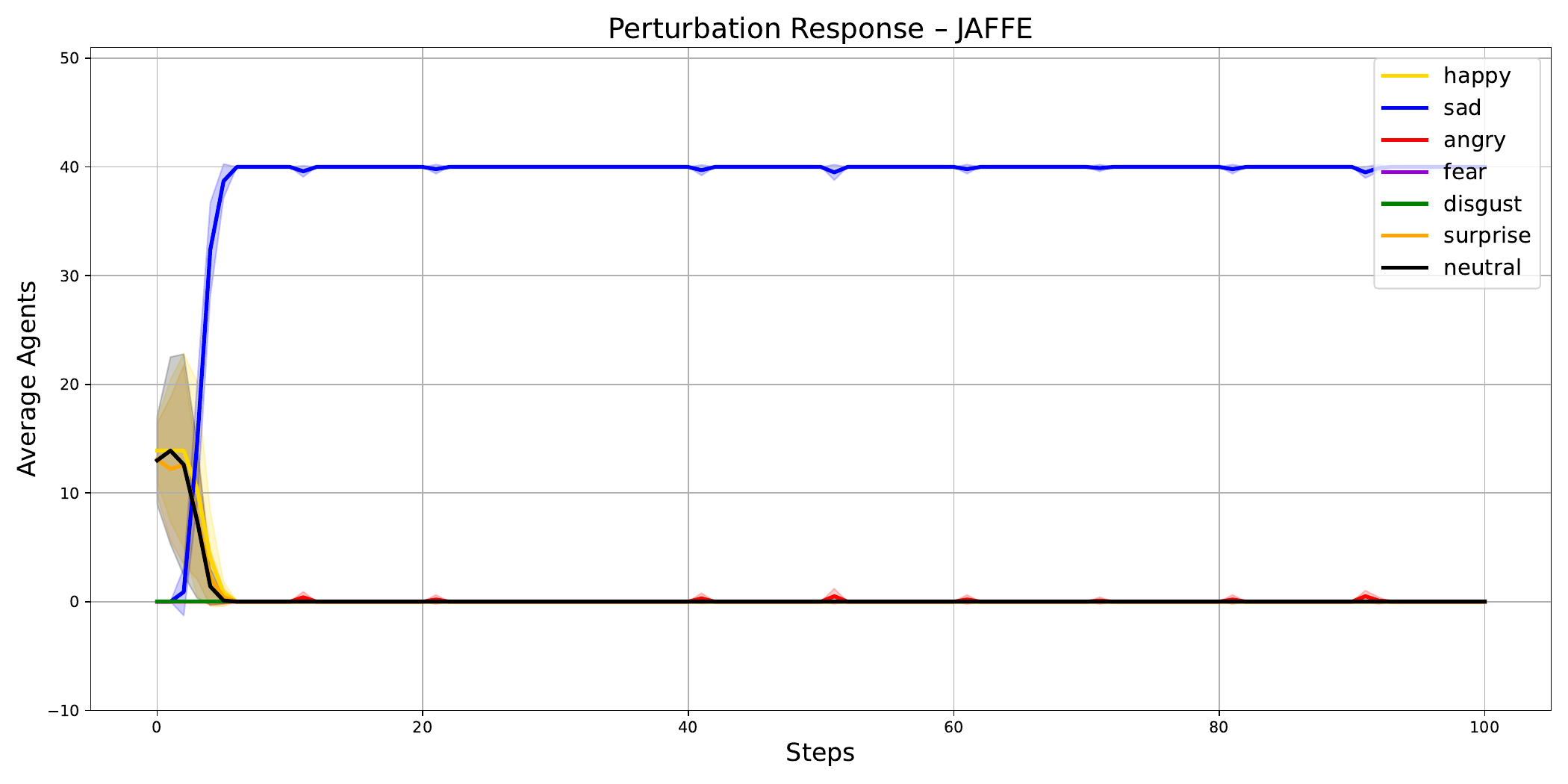}
    \caption{Emotional resilience of homogeneous JAFFE agents under repeated emotional shocks. Each curve shows the decline of positive emotion ratio over time.}
    \label{fig:exp3_jaffe}
\end{figure}

\begin{figure}[ht]
    \centering
    \includegraphics[width=\columnwidth]{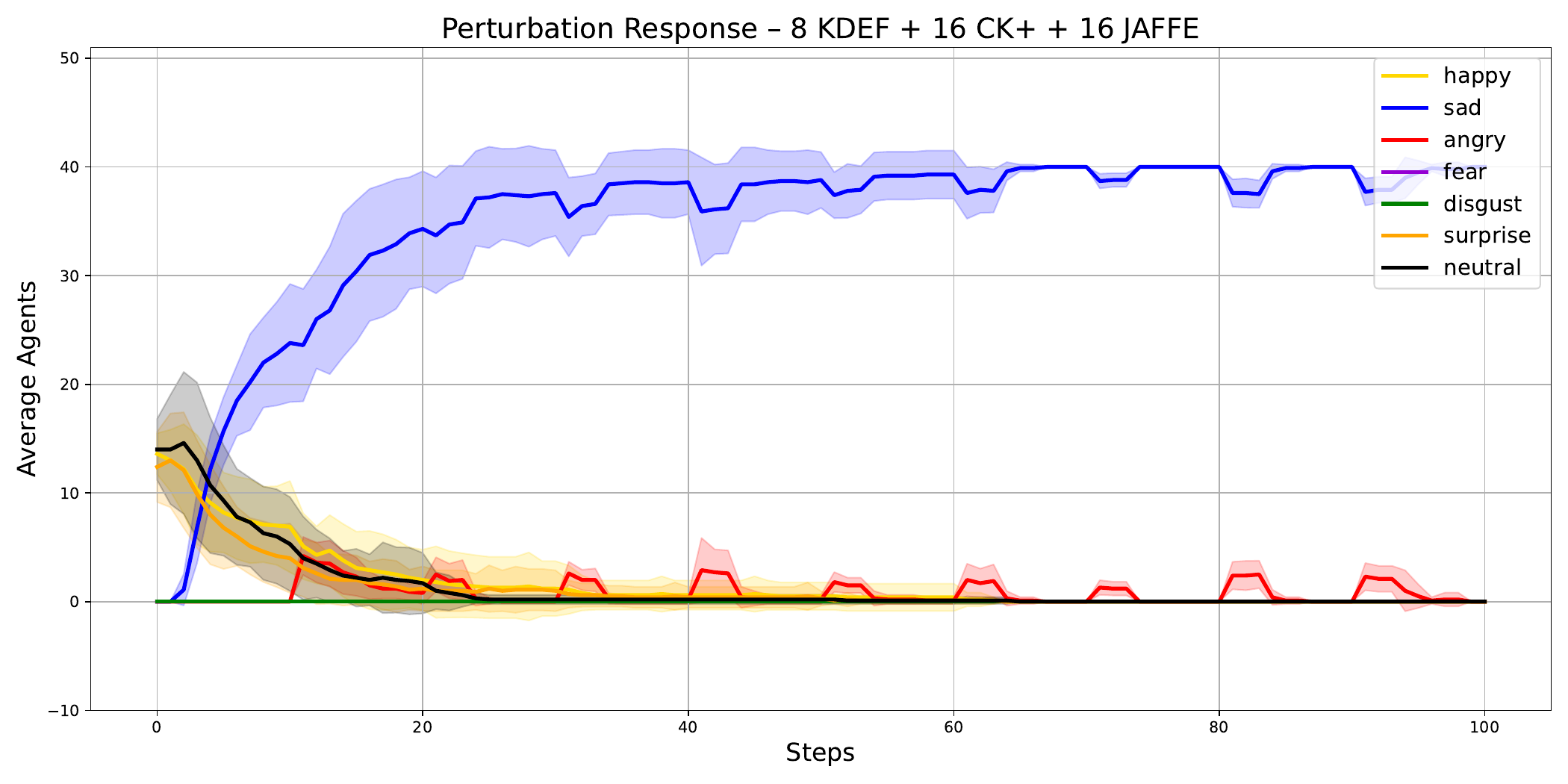}
    \caption{Emotional stability under repeated negative shocks in a mixed population of 8 KDEF, 16 CK+, and 16 JAFFE agents. The decline in positive emotion varies depending on the classifier composition.}
    \label{fig:exp3_mixed_skewed}
\end{figure}

\begin{figure}[ht]
    \centering
    \includegraphics[width=\columnwidth]{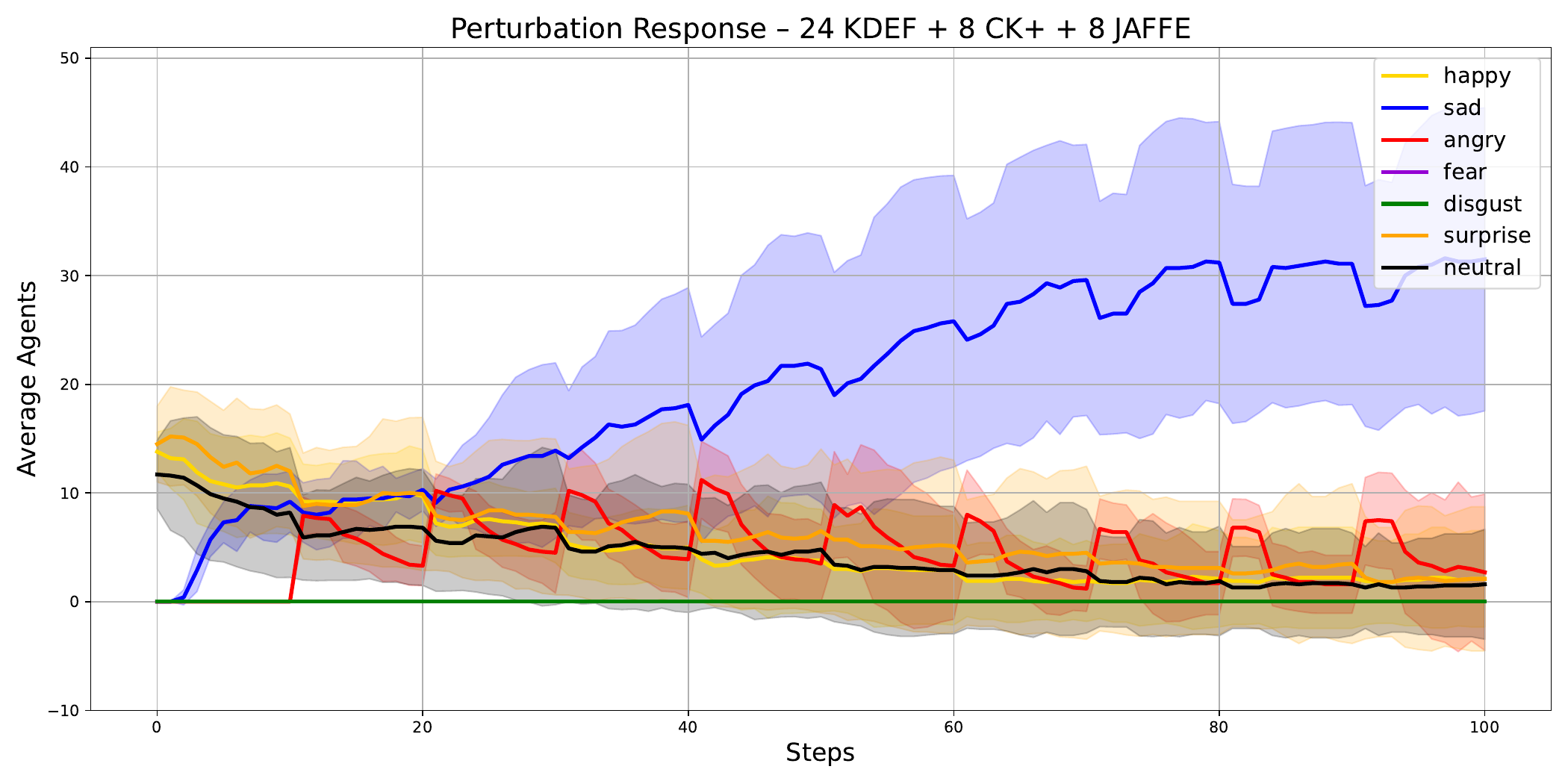}
    \caption{Emotional resilience in a classifier-dominant scenario with 24 KDEF, 8 CK+, and 8 JAFFE agents. Higher presence of KDEF-trained agents buffers the emotional decline under repeated shocks.}
    \label{fig:exp3_mixed_dominant}
\end{figure}

\subsection{Experiment 3: Emotional Resilience under Perturbation}

Finally, we evaluate the system's resilience to repeated external emotional perturbations. In this setting, agents begin in randomly sampled emotional states, and at every ten simulation steps, 20\% of the population is forcibly assigned a negative emotion. The periodic assignment of negative emotions to 20\% of the agents serves as an exogenous shock to probe system resilience. This mechanism is intended as a proxy for external stressors (e.g., crises or hostile signals) and should not be interpreted as a literal psychological process. We explore homogeneous and heterogeneous populations to check the impact of perceptual fidelity on recovery from emotional states and long-term stability.

\paragraph{Homogeneous Populations.}  
In purely homogeneous settings, we observe significant differences in emotional resilience across classifiers. Populations composed entirely of KDEF agents display a gradual but incomplete erosion of positive affect (see Figure~\ref{fig:exp3_kdef}). While the proportion of positive emotions declines from 1.0 at step 10 to 0.08 at step 100, a small core of positivity persists. This is consistent with the classifier's strong perceptual accuracy and high final trust level (0.933), which appear to slow the spread of negative emotion and maintain partial recovery between shocks.

In contrast, CK+ agents degrade more rapidly  (see Figure~\ref{fig:exp3_ck}). Although they begin with a similarly high proportion of positive affect (0.93 at step 10), their emotional state deteriorates faster, reaching zero positive emotions by step 70. Their final trust score is extremely low (0.034), indicating that inaccurate perception undermines agents' confidence in others, thereby accelerating emotional collapse.

JAFFE agents are unable to resist perturbation at all  (see Figure~\ref{fig:exp3_jaffe}). Even at step 10, no positive emotions remain in the system, and trust plummets to near-zero levels (0.001). This result underscores the fragility of populations reliant on poor perception: once injected, negativity enters and becomes dominant and irreversible.

\paragraph{Heterogeneous Populations.}  
Mixed-agent populations show outcomes that depend strongly on the relative proportions of perceptual capacity. In the setting with \textbf{8 KDEF}, \textbf{16 CK+}, and \textbf{16 JAFFE} agents, the emotional trajectory mirrors the worst-performing components  (see Figure~\ref{fig:exp3_mixed_skewed}). Positive affect drops sharply from 0.41 at step 10 to zero by step 70, with final trust scores confirming this collapse (JAFFE: 0.005, CK+: 0.039, KDEF: 0.961). The small number of accurate agents is insufficient to stabilize the group in the presence of overwhelming misperception.

In contrast, the configuration with \textbf{24 KDEF}, \textbf{8 CK+}, and \textbf{8 JAFFE} agents retains substantially more emotional stability  (see Figure~\ref{fig:exp3_mixed_dominant}). Positive emotions remain at 0.15 by step 100, and trust in the KDEF agents remains high (0.956). This demonstrates that when perceptually reliable agents form a substantial majority, emotional stability can be partially preserved despite repeated perturbations.

These findings show that emotional resilience depends not only on perceptual accuracy but also on its distribution: a few reliable agents cannot offset widespread misperception, whereas a majority can stabilize the population under stress.


\section{Conclusions}

This study demonstrates that the quality of emotion perception critically shapes emergent social dynamics in agent-based simulations. Populations with high-accuracy classifiers maintained trust, emotional balance, and stable clusters of positive states. In contrast, those with poor classifiers rapidly converged toward negative affect (especially sadness), producing fragmented structures and minimal trust. Even small proportions of inaccurate agents destabilized otherwise cohesive societies, showing how perceptual biases (whether from data, culture, or algorithms) spread through local interactions and undermine collective coherence. Under repeated emotional shocks, heterogeneous populations deteriorated more rapidly than homogeneous ones, with only those dominated by accurate classifiers retaining stability. More broadly, our results resonate with Schelling's classic segregation model~\citep{schelling1971}: just as mild preferences can yield large-scale segregation, here systematic misperception drives avoidance and trust decay that accumulate into strong emotional clustering and social fragmentation. Beyond simulation, biased emotion classifiers may erode trust in human–AI interaction, underscoring the need for reliable perception in social technologies.

\textbf{Acknowledgements}.
This work is partially funded by project PID2021-122402OB-C22/MICIU/AEI
/10.13039/501100011033 FEDER, UE and by the ACIISI-Gobierno de Canarias and European FEDER funds under project ULPGC Facilities Net and Grant \mbox{EIS 2021 04}. 

\footnotesize
\bibliographystyle{apalike}

\end{document}